\documentclass[conference]{IEEEtran}
\IEEEoverridecommandlockouts

\usepackage{hyperref}
\hypersetup{colorlinks,urlcolor=blue, citecolor=blue}
\usepackage{url}
\usepackage{cite}
\usepackage{enumitem}
\usepackage{xspace}
\usepackage{multirow}

\usepackage{listings}

\def\eg{\emph{e.g.}\xspace}

\def\ie{\emph{i.e.}\xspace}

\usepackage{cite}
\usepackage{amsmath,amssymb,amsfonts}
\usepackage{algorithmic}
\usepackage{graphicx}
\usepackage{textcomp}
\usepackage{xcolor}
\def\BibTeX{{\rm B\kern-.05em{\sc i\kern-.025em b}\kern-.08em
T\kern-.1667em\lower.7ex\hbox{E}\kern-.125emX}}
    
\begin{document}

\title{RAI4IoE: Responsible AI for \\Enabling the Internet of Energy}

\author{\IEEEauthorblockN{Minhui Xue\IEEEauthorrefmark{1}, Surya Nepal\IEEEauthorrefmark{1}, Ling Liu\IEEEauthorrefmark{2}, Subbu Sethuvenkatraman\IEEEauthorrefmark{4}, \\Xingliang Yuan\IEEEauthorrefmark{3}, Carsten Rudolph\IEEEauthorrefmark{3}, Ruoxi Sun\IEEEauthorrefmark{1}, and Greg Eisenhauer\IEEEauthorrefmark{2}
}

\IEEEauthorblockA{\IEEEauthorrefmark{1}CSIRO's Data61, Australia}
\IEEEauthorblockA{\IEEEauthorrefmark{4}CSIRO's Energy, Australia}
\IEEEauthorblockA{\IEEEauthorrefmark{2}Georgia Institute of Technology, USA}
\IEEEauthorblockA{\IEEEauthorrefmark{3}Monash University, Australia}
}

\pagestyle{plain}

\maketitle

\begin{abstract}

This paper plans to develop an Equitable and Responsible AI framework with enabling techniques and algorithms for the Internet of Energy (IoE), in short, RAI4IoE. The energy sector is going through substantial changes fueled by two key drivers: building a zero-carbon energy sector and the digital transformation of the energy infrastructure. We expect to see the convergence of these two drivers resulting in the IoE, where renewable distributed energy resources (DERs), such as electric cars, storage batteries, wind turbines and photovoltaics (PV), can be connected and integrated for reliable energy distribution by leveraging advanced 5G-6G networks and AI technology. This allows DER owners as prosumers to participate in the energy market and derive economic incentives. DERs are inherently asset-driven and face equitable challenges (\ie, fair, diverse and inclusive). Without equitable access,  privileged individuals, groups and organizations can participate and benefit at the cost of disadvantaged groups. The real-time management of DER resources not only  brings out the equity problem to the IoE, it also collects highly sensitive location, time, activity dependent data, which requires to be handled responsibly (\eg, privacy, security and safety), for AI-enhanced predictions, optimization and prioritization services, and automated management of flexible resources. The vision of our project is to ensure equitable participation of the community members and responsible use of their data in IoE so that it could reap the benefits of advances in AI to provide safe, reliable and sustainable energy services. 
\end{abstract}


\section{Introduction}

Energy as a service market is projected to reach \$112.7 billion by 2030, growing at a CAGR of 7.6\% from 2021 to 2030. Renewable energy market is projected to reach \$103.2 billion by 2030, growing at a CAGR of 27.2\% from 2021 to 2030~\cite{allied2022renewable,future2022}. 
With the overarching goal of reaching net zero by 2050 mission~\cite{AuEmissionsPlan,USEmissionsPlan}, the energy sector is going through drastic changes, especially fueled by the advancement of technologies in electric vehicles, storage batteries, wind turbines and photovoltaics (PV), and the emergence of Internet enabled distributed energy resources (DERs). For example, over 30 GW of renewable energy and storage capacity was installed in the United States in 2020. PV represented approximately 40\% of new U.S. electric generation capacity in 2021, compared to 4\% in 2010. Eight of the leading PV markets collectively installed 93 GW of PV in 2020, up from 69 GW in 2019~\cite{nrel_gov}. 
The US Energy Information Administration (EIA) predicts that electric vehicles (EVs) will grow from 0.7\% of the global light-duty vehicle (LDV) fleet (cars and trucks) in 2020 to 31\% in 2050~\cite{electrek}.
The Australian Energy Network Association's Electricity Network Transformation Roadmap (ENTR) projects that more than 40\% of energy customers will use DERs by 2027, and more than 60\% by 2050. The next generation energy infrastructure needs to support these DERs by applying and augmenting Artificial Intelligence (AI) advances to address DER challenges, giving birth to a new field of research in the energy sector – the Internet of Energy (IoE). 

\begin{figure*}[t]
\centering
\includegraphics[width=\linewidth]{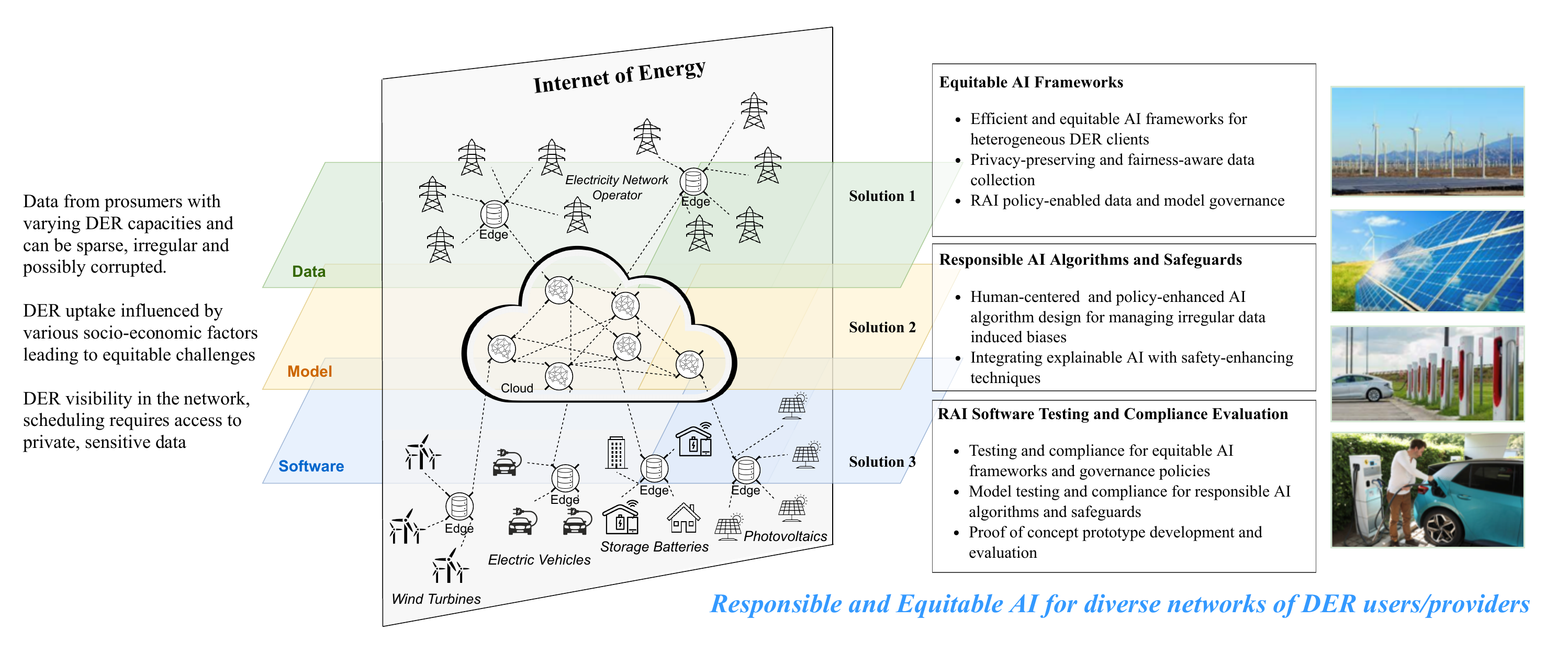}
\caption{{\small Overview of RAI4IoE, enabling Responsible AI for the Internet of Energy.} }
\label{fig_overview}
\end{figure*}

\noindent \textbf{DER Use Cases and Requirements for AI-enabled IoE.~}
The penetration of EVs in the market is rapidly increasing. The net-zero mission~\cite{AuEmissionsPlan} will have a huge demand for (i) ubiquitous deployment of EVs; (ii) smart placement and intelligent network of EV charging stations for the convenience of EV drivers and the decentralized grid storage; and (iii) a decentralized EV infrastructure that is capable of providing efficient and responsible electricity network services for the growing number of EV users and the aggregators. AI technology has been used to deploy the right amount of charging stations in the best location possible~\cite{asensio2017effectiveness,asensio2019correcting}. In addition to charge electric cars, EV stations can also send power from vehicle batteries to the local utility grid for use by homes and businesses (vehicle to grid)~\cite{dumiak2022road}. Big automakers have increased the production of EVs with bidirectional chargers, enabling owners of EVs to power their houses. Bidirectional charging works efficiently for EVs with predictable movements, \eg, daily commute between home and office during work days, or parking EVs at home during an overseas trip. AI technology can be used in such cases to predict and program when to charge and discharge an EV's battery. More importantly, geographically distributed EV users and EV charging stations will form different DER communities with different demographic features. These networks of EV charging stations have the potential to collectively serve as an efficient decentralized power grid storage for improving the peak-hour power efficiency, enabling an AI-enhanced IoE for everyone. However, the AI-enhanced EV infrastructure needs to train the AI models for managing and accessing EV battery resources, such as smart placements of charging stations,  smart modeling of EV drivers' movement patterns through monitoring and learning their commuting habits. To ensure that the AI models provide equitable access to all DER user/provider communities regardless their sizes and demographics, the AI model training for DER systems requires sufficiently representative data, \eg, capturing time-sensitive travel data from every group in the society. This can be challenging as city dwellers' traveling habits are different to country people, and a tourist's traveling behaviors would be different to local residents. In addition, when using the data collected from the EVs owned by certain groups in the society to build the AI models for deploying charging stations and accessing DERs in peak energy demand, such AI models may introduce biases and discriminate other groups in the society. We argue that the net-zero mission is only possible through responsible and equitable AI, ensuring the equal participation of all groups in society. 

Photovoltaics, a method of using solar cells to convert energy from the sun into a flow of electrons by the photovoltaic effect, can be used to power equipment and recharge a battery. The roof-top solar panels are crucial components of the grid-connected PV systems, ranging from small residential and commercial roof-top systems. High penetration of PV with or without stroage solutions in residential, commercial and industrial sites is enabling participation of PV owners in the energy market leading to effective uilisation of their resources and deriving economic benefits. Real time balancing of supply and demand is requried to faciliate PV owners to participate in the local energy market. AI Technology has been used to forecast load demands and manage the energy produced by solar panels, utilizing different energy storage solutions~\cite{mohamad2021optimum}. This requires developing the AI models trained with residential and commercial energy data or business activity-specific energy data. For household energy data, it may indicate time-dependent energy-related social activities, such as cooking, cleaning, and showering. For business activity-specific energy data, it may indicate the type of business operations performed based on the energy consumption patterns in both spatial and temporal dimensions. Hence, privacy preserving AI techniques are critical for managing, accessing and analyzing such data to protect both individual users' personal privacy and organizational privacy regarding their business activities. Furthermore, the AI models trained using data contributed mainly by a certain group of society, \eg, a privileged group or neighborhood with a large installment of roof-top solar panels, may not work equally well for the underprivileged groups, and the embedded biases towards underprivileged groups will be propagated and possibly amplified in the net zero solutions. 

\noindent \textbf{The Vision of AI-enhanced IoE.~}
Electricity systems across the globe are undergoing a significant transition due to decarbonization, decentralization and digitalization. Increasing penetration of variable renewable resources such as solar and wind associated with retirement of fossil fuel based plants is leading to more non controllable,  variable power generation in the mix. At the distribution level, DER systems such as rooftop PV, battery and EVs are introducing new challenges due to two way power flows and lack of visibility about these resources.  While decarbonization and decentralization are leading to new electricity grid challenges, digitalization enabled by AI-enhanced IoE is opening new opportunities to address these challenges. For example, rapid advances in AI and the increased availability of data through ubiquitous network connectivity offer a timely opportunity for AI-enabled IoE. Digitalization is helping improve the reliability and security of the grid with high levels of renewable and DER sources, and it also underpins operation of digital grid where DER owners as prosumers can participate in the electricity market while delivering grid balancing services. 

We envision that the ultimate goal of AI-enabled IoE is to support an ``equitable and responsible transition'' towards net zero. By \textit{equitable}, we mean that one group/individual in society will not be disadvantaged over others. All individuals and groups in society have equal opportunity and participation in the AI-enabled IoE across geographically distributed DER communities. By \textit{responsible}, we mean that the distributed architecture and algorithms designed to learn and create the AI models for DER users should not only provide an equal opportunity for everyone to participate in the electricity market but also ensure all individuals and all DER communities have efficient, secure, privacy preserving and safe access to the DER market~\cite{xuan2022equality}.  
To realize this overarching vision, we propose this study to develop responsible and equitable AI frameworks, algorithms, and compliance methods towards enabling the IoE, coined as RAI4IoE. Figure~\ref{fig_overview} gives an overview of the RAI4IoE for managing and accessing DERs through responsible AI frameworks and algorithms. This is accomplished while addressing challenges and proposing potential solutions across three distinct levels, \ie, data-level responsible governance, model-level distributed learning, and software-level AI framework testing and evaluation. We summarize the key contribution of this study as following.

\begin{itemize}
\item The study undertakes a meticulous review of the overarching vision behind the integration of the Internet of Energy and responsible AI. It delves into the conceptual underpinnings, highlighting the synergies between these domains and emphasizing their potential to revolutionize energy management and consumption patterns.
\item The study undertakes a meticulous review of the overarching vision behind the integration of the Internet of Energy and responsible AI. It delves into the conceptual underpinnings, highlighting the synergies between these domains and emphasizing their potential to revolutionize energy management and consumption patterns.
\item Addressing the identified challenges head-on, the study puts forth a series of innovative and actionable solutions. By offering concrete strategies for data governance, safeguard algorithm design, and system evaluation, the study equips stakeholders with a roadmap to foster a harmonious coexistence between cutting-edge technology and sustainable energy practices.
\end{itemize}

\section{Related Work}

In this section, we introduce the recent studies related to the domain of enabling the IoE with responsible AI.


\noindent \textbf{Data-level responsible governance.~}
There has been a limited amount of research on creating a data governance structure for ML datasets. Recent research has begun to analyze and unpack what is represented in ML datasets~\cite{birhane2022values,denton2021genealogy,paullada2021data,dodge2021documenting,liao2021we,scheuerman2021datasets,sun2023not}, and dataset lifecycle~\cite{hutchinson2021towards}. They scratch the surface of what responsible data systems may look like in ML.
Mainstream ML research has focused predominantly on improvements to the model architecture, training procedure, and hypeparameters~\cite{paullada2021data,sambasivan2021everyone} for trustworthy decision making, including analyzing various dimensions of data quality and stewardship~\cite{peng2021mitigating,prabhu2020large,rogers2021just,sambasivan2021everyone,bandy2021addressing,birhane2021multimodal,kreutzer2022quality,dodge2021documenting,gehrmann2021gem,wang2021adversarial,cao2023stylefool,hu2022mi}. These efforts have gone hand-in-hand with efforts centered around values of transparency and replicability in scientific work, through the standards and conference checklists~\cite{dodge2019show,pineau2021improving}. Newer checklists for ``Responsible'' data use~\cite{rogers2021just} include asserting the importance of respecting values, \eg, non-discrimination (fairness), consent, or privacy in the development and use of datasets. 
Still, comparatively little attention is paid to responsible data and model governance, which is instrumental for accessing and planning DERs in IoE scenarios.

\noindent \textbf{Model-level distributed learning and Safety-enhancing for DER users.~}
A unique advantage of flexible electricity market is to leverage the geographically decentralized networks of DER users/providers. 
Instead of centralizing the data from disparate DER clients, distributed learning algorithms can be developed to train a global DER-specific monitoring and analytic model for a community of DER users/providers~\cite{fedavg,Lim+2020}. However, conventional approaches in the literature assume that participants have sufficient resources to run a full precision model provided by the federated server~\cite{fedscale}, including  
horizontal Federated Learning (FL), represented by FedAVG~\cite{fedavg}, vertical FL, represented by SCAFFOLD~\cite{karimireddy2020scaffold} and HeteroFL~\cite{heterofl}, and efficient FL with reduced global model size through model reduction using neural network compression~\cite{NNcompression} or neural network pruning such as weights or feature map binarization~\cite{Weight-pruning,Ling-infocom} or lottery tickets~\cite{Lottery}. As a result, those edge clients with limited resources, who cannot host the full size model at run-time, will not be admitted (unqualified) to participate in distributed training of a global model. 
We argue that for a community of DER users/providers, developing an equitable distributed AI framework should be the first principle, enabling equitable access to DERs regardless which types of users and how limited their assets (\eg, computation resources) might be.

Data corruption can drastically degrade the performance and efficiency of 
both model learning during training phase and model prediction during deployment phase.
Data \emph{poisoning attacks}, also called Trojan attacks, are adversarial data corruptions during model training~\cite{rubinstein2009antidote,biggio2012poisoning,biggio2013poisoning,xiao2015feature,Jagielski18,poisoningattackRecSys16,YangRecSys17,Suciu18,shafahi2018poison,ji2018model,cheu2019manipulation}. Stealthy data poisoning is especially harmful to distributed learning systems, such as the network of DER nodes, because the global model can reach target accuracy while model prediction for certain group/class of entities may fail miserably. 
Data poisoning can be classified into clean label poisoning and dirty label poison (\eg, \emph{backdoor attacks}~\cite{gu2017badnets,liu2017trojaning,chen2017targeted,wang2019neural,Bagdasaryan18,xie2019dba,li2020invisible,doan2021tnt,li2021hidden,ma2023beatrix}). 
We note that the previous representation-based poisoning or backdoor detection methods~\cite{chen2019detecting, tran2018spectral,tang2021demon, hayase2021spectre} ignore the explainability information and only consider the first moment (mean) discrepancy between clean and trojaned samples. However, the simplification reduces the discriminating power of the methods against more complex attacks such as dynamic backdoors. 

\noindent \textbf{Software-level AI framework testing.~}
Although software testing is a mandatory component for  traditional computer programs~\cite{sun2023mate,sun2021empirical,RuoxiSenSys_20,ma2018deepgauge,xie2019deephunter,ma2018deepmutation,zhang2022path,feng2021snipuzz}, AI models are primarily measured by the test accuracy over the test data. We see three serious drawbacks inherent in such AI software testing. First, due to limited availability of well-defined high quality test data, good accuracy performance on one set of test data is insufficient for high confidence evaluation of the generality of AI systems. Second, in contrast to traditional software systems that have explainable and controllable logic and functionality, the lack of interpretability for the prediction output from an AI model makes it extremely hard to perform the robustness analysis of the generalization performance and the understanding of the sources of errors. Third, there are little study on how to measure and test the fairness of an AI prediction output in terms of biases in both data and  algorithm. With the recent surge of attention on AI fairness, there are numerous surveys on AI fairness~\cite{mehrabi2021survey,pessach2022review,chen2022fairness,hort2022bia,hutchinson201950}. 
However, fairness testing has received limited attention. DeepInspect~\cite{tian2020testing} is a white-box fairness measurement for DNNs. Fair-SMOTE~\cite{chakraborty2021bias} is designed to detect biased data labels and data distributions based on supervised training. FINS~\cite{cachel2022fins} proposed the group fairness testing for a subset of selection tasks.
A survey~\cite{zhang2020machine} provides a general overview of ML testing from testing workflow and testing component perspective, in which fairness was considered as one of the many testing properties. 
From the testing workflow angle, the testing activities may include test oracle identification and test input generation. From the testing component angle, software systems typically rely on AI model to make predictions without additional verification or inspection. Such AI testing cannot detect if fairness bugs  exist and whether bias is in the data collection and interpretation or in the prediction algorithms.

\section{Future Challenges and Possible Solutions}

Most of the DER assets are relatively small (10s to 100s of kilowatts) compared with utility scale assets and dispersed across different owners.
Accessing these flexible DER resources for market participation often requires accessing private information of individual owners. Participation of these sources in the market will require, i) tools for real-time, locational ‘situation awareness’ of these sources and forecasting of flexible demand, including resource availability and quantity; ii)	aggregating demand from individual sources for participation and providing capability to automate remote dispatch of flexible demand resources. To facilitate equitable participation of all DER owners and users in the automated flexibility market, AI enabled IOE should be governed by the responsible AI (RAI) frameworks and guidelines. This RAI governance framework should create and develop AI system architectures and algorithms for distributed monitoring, scheduling,  management, and consumption of DERs, while exercising and guaranteeing responsible and equitable AI through ensuring AI fairness and safeguarding AI privacy and security in an open and continuously evolving IoE ecosystem. We envision the following challenges in enabling responsible AI for IoE:

\begin{itemize}[leftmargin=*]
\item \textbf{Challenge I (data-level)~}
The first challenge is to develop the RAI frameworks and algorithms that \textit{enable fair and equitable access to DERs}~\cite{equitableAIChallenge,vincent2019options}. For example, we need to identify vulnerabilities and weak spots in fair governance of DERs. This includes promoting equitable participation for all DER users/providers, addressing RAI challenges specific to DERs, such as lack of transparency, limited number of actors, price manipulation, data biases and algorithm biases. Furthermore, we argue that enabling responsible and equitable AI in the IoE also implies adequate support for both communities with limited DER and non-DER communities. Although AI fairness and AI ethics are gaining much attention in recent years, we argue that the above energy-domain specific DER sharing challenges have not be systematically addressed in the context of distributed AI and edge analytic systems and services.
\item \textbf{Challenge II (model-level)~} Another AI safety challenge is to develop the RAI framework and algorithms that \textit{support privacy preserving and secure access, monitoring, and management of the networks of DERs} from diverse energy sources. This requires an in-depth understanding of the inherent risks in the management of disparate DER sources through the networks of DER users/providers. As per the two DER use-cases (PV and EV), efficient sharing of DERs requires distributed and collaborative learning of global DER models for pricing mediation, DER availability monitoring, and equal accessibility management based on sensitive data of DER users, such as location and travel, commute patterns between work and home, as well as other sensitive user information. Due to the open nature of DER networks, some DER edge clients may be compromised. Hence, curious or malicious agents from compromised clients may engage in different types of fraudulent activities, such as phishing, price pump and dump, de-anonymization of private user accounts, and Trojan attack for adaptive data poisoning and model poisoning. Although AI privacy has been extensively studied in the literature, with differential privacy (DP) as the key technology, it is widely recognized that DP solutions to date cannot detect or prevent fraudulent transactions in decentralized finance, and are not resilient to both gradient inversion/leakage attacks~\cite{wei2021gradient,wei2021gradient1} and Trojan attacks~\cite{tolpegin2020data,truex2019effects}. In addition, most of AI security solution is attack-specific: countermeasures for Trojan detection and defenses are not applicable to detect or mitigate gradient leakage attack in distributed learning or fraudulent events. 
We plan to develop the RAI framework and algorithms that establish DER-specific privacy and security governance guidelines and exercise privacy and robustness enhancing techniques to protect DER transactions against disruptive events and unauthorized data disclosure for all DER users/providers.
\item \textbf{Challenge III (software-level)~} The proliferation of DERs creates the challenge of \textit{managing the distribution network of DER users/providers to provide reliable, resilient and sustainable energy services}, where every individual in the network could be a resource consumer and producer (prosumer). When applying and augmenting AI technology to tackle this challenge, we need to address a fundamental question: can we develop responsible AI methodology that ensures equitable access and resilience for everyone to use and benefit from the DER services? 
\end{itemize}

This research aims to tackle these challenges by developing responsible AI and equitable data governance frameworks, algorithms, and compliance methods towards enabling the IoE with three innovative possible solutions.

\subsection{Equitable AI Frameworks for Distributed Learning and Data Governance}\label{sec_data_governance}
The first solution will focus on identifying and tackling the problems in existing distributed learning frameworks and data collection methods, which may lead to the possible inequality in the IoE such as those discussed in DER use-cases. We plan to develop a responsible and equitable distributed learning framework with heterogeneous clients,  enabling the DER users with limited computing resources to have equal opportunities to contribute to the AI model learning. We will integrate data governance with privacy-preserving and fairness-aware data collection, minimizing the adverse effect of biases in data sampling and data annotation on the training and inference of AI models, and boosting the robustness of AI model training and model inference against adversarial and systemic disruptions.  

\begin{figure}[t]
\centering
\includegraphics[width=\linewidth]{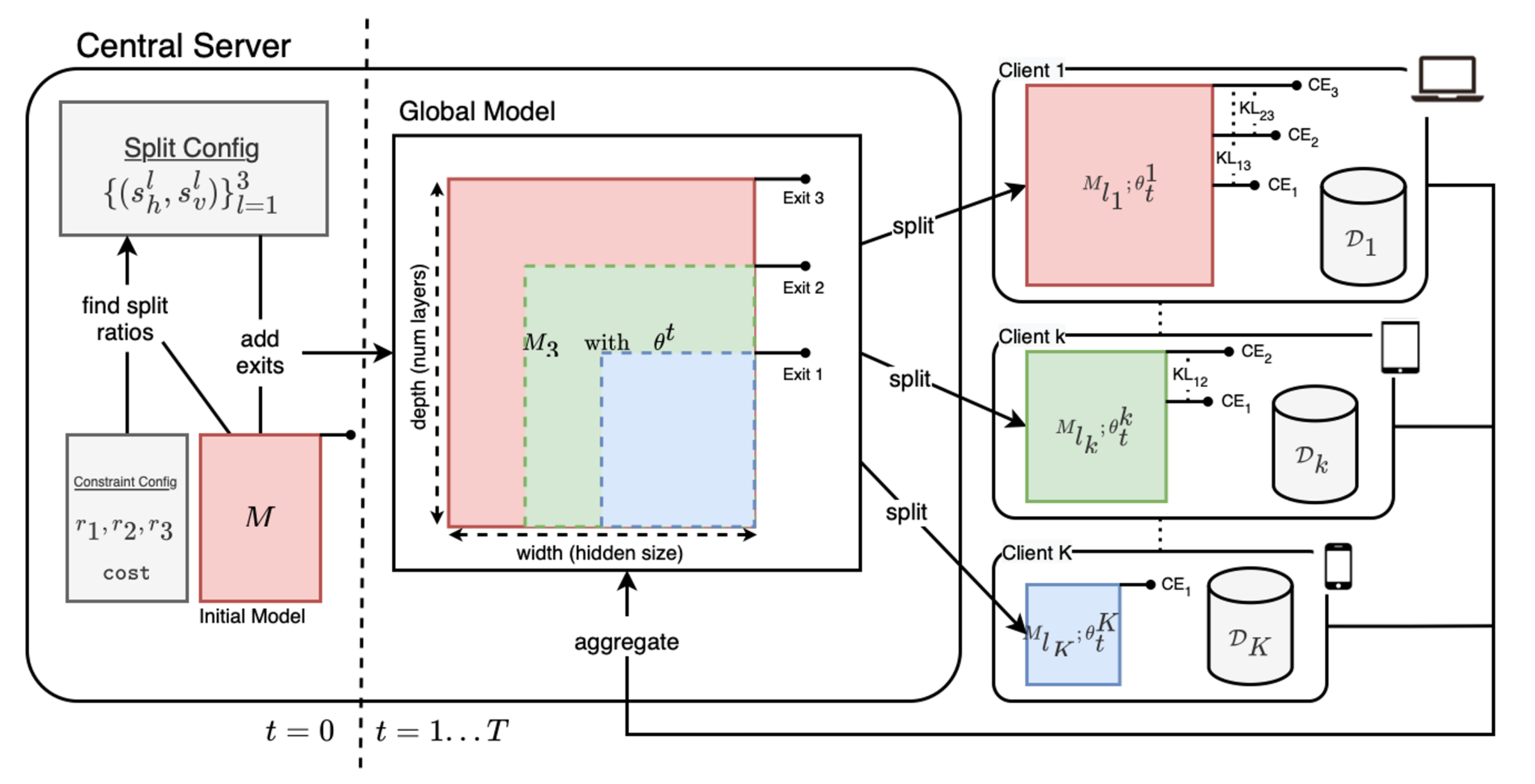}
\caption{{\small Enabling an Equitable AI framework for distributed learning.}}
\label{fig_equitableFL}
\end{figure}

\noindent \textbf{Efficient and Equitable Distributed Learning for Heterogeneous DER Users.~}
To enable distributed learning for heterogeneous DER users/providers with limited computing resources, we propose to develop a novel equitable distributed AI framework that can adaptively scale down the global deep neural network (DNN) model based on each client's computational constraint. 
One approach is to uniformly split the model along width and depth dimensions, enabling heterogeneous DER clients to participate federated learning by selecting the best-fit model partitions and performing resource aware local model training (see Figure~\ref{fig_equitableFL}). Our idea of here is based on the understanding that a deeper model is more capable of extracting higher-order and complex features, while a wider model has access to a larger variety of lower-order and basic features. 
The advancement in DNN algorithm optimizations~\cite{efficientnet,Li+2020,RecursiveDeepModels2013,SplitFed,Transformer} has also demonstrated the importance of balancing the size of different dimensions while scaling a neural network to big data. Although some existing work~\cite{heterofl,feddf} studied the feasibility of splitting a DNN model along only the width dimension, aiming to build a model from training data with only partial features, we observe that performing model size reduction only along the width dimension causes unbalance and possible algorithmic biases in the learning capabilities of the resulting model~\cite{msdnet,branchynet,adaptive_laskar,improved_adaptive,distil_exit,pabe,fastbert}. 

The first novelty of this framework is to develop early-exit based NN split method along the depth dimension by injecting multiple early exit learners to the global model at different NN layers based on the neural architecture and computational constraints.
In addition to balance the preservation of both complex and basic features, we plan to optimize our adaptive depth and width scaling down approach by employing self-distillation among the exit predictions, during the training of local models, to improve the knowledge transfer among subnetworks. Knowledge distillation is the process of transferring knowledge from a large (teacher) network to a smaller (student) network by training the student network using teacher predictions as soft labels~\cite{know_dist}. Self-distillation is a form of knowledge distillation, where the same network is used as both teacher and the student to improve performance and regularization during training, especially for multi-exit models~\cite{selfdist,ownteacher,improved_adaptive,distil_exit}. In particular, we plan to optimize these models with the additional objective of minimizing the KL divergence among the early exit (student) and final (teacher) predictions, which will provide effective aggregation through increasing knowledge flow among local models. 
Our proposed resource-adaptive federated learning framework addresses the equitable access to distributed learning from two dimensions.  First, we perform adaptive scaling down the model based on the computational constraints of participating clients by uniformly splitting the model along width (hidden size) and depth (number of layers) dimensions by jointly training multiple early-exit capable subnetworks. The goal is to preserve the important balance of access to both basic and complex features in local models (subnetworks of different complexity). Second, we plan to employ self-distillation among early exit predictions and final predictions during the training of local models to provide effective aggregation. We also plan to explore other alternative optimization opportunities jointly with the other components in this framework to incorporate privacy and fairness enhanced techniques for data and model governance and for improving the aggregation robustness.

\noindent \textbf{Human-centered and Policy-enhanced Data and Model Governance.~}
A technical challenge in designing a responsible data governance framework is to provide a proper balance between protecting the data and making it accessible to the DER users/providers. 
In contrast to data governance, the model governance framework will safeguard the AI model development and AI model deployment simultaneously. First, the model governance needs to validate and assure the design of the backbone algorithms used for AI model training, such as the choice of loss optimization, or regularization estimators, will not introduce new security and privacy risks, or new sources of biases. Second, the model governance framework should also include the compliance checking for the robust version or private version or unbiased version of the AI algorithms. 
Finally, a RAI model governance framework should incorporate minimal data access techniques, allowing AI models to unlearn under privacy regulations like GDPR or limiting the data sharing to be compliant with certain protection regulations.

A big issue for data governance in IoE is that most data are unique in context and difficult to anonymize~\cite{chaski2012best,kumar2017army,mozes2021no}, and they could identify people through de-anonymization and inference based reconstruction attacks.
Privacy legislation such as European GDPR~\cite{EU:16} and Australian Consumer Data Right (CDR)~\cite{link1, link2} for Energy requires that data be used with (revocable) consent, and data subjects should also have the right to delete or rectify existing records (\ie, the ‘right to be forgotten’). We argue that a responsible framework for data and model governance should consider not only limited access by need to know practice, whether and how to remove specific instances from its datasets, but also how to minimize the risk of privacy leakage, Trojan attack, or biases in data and algorithms.
One of the outcome from our data and model governance framework is to support the DER users' rights by allowing marginalized populations better control over how they are represented in the data used to train ML models in an effort to lessen algorithmic discrimination, including the support for auditability to promote accountability. 

\subsection{Responsible AI Algorithms and Safeguards for IoE-inspired Distributed Learning}
This solution will address robustness and safety issues manifested by the AI algorithms used for distributed and collaborative learning of predictive analytic models over the IoE networks, including evasion attacks to fool the AI models used during edge deployment for predictive inference by DER users. The outcome includes new algorithms and optimizations for efficient management of DERs in the presence of biases due to irregular data, Trojan attack when a small percentage of DER clients are compromised. 
Particularly, we propose integrating explainable AI with safety-enhancing techniques.

Feedback-driven AI promotes human-centric design and human-in-the-loop intelligence by generating feedback based on domain knowledge and the explainability of AI results~\cite{du2019techniques}. Explainability is typically modeled by a set of probabilistic logical formulas summarizing human expert knowledge about causality. A popular approach to explainable AI is to learn an effective surrogate model, which is based on domain knowledge from human experts and hence is intuitive for humans to understand, yet can best approximate the original complex model. In addition, to optimize the primary loss function by minimizing the difference between the surrogate model and the original model, the explainability learning algorithm will add a regularization term to reward the explainability. Different criteria can be used to measure explainability, such as the number of probabilistic logical formulas (the smaller the better). Explainability can zoom into the causal relationships among variables of interest and reveal how the fundamental design principles and key parameters may impact the convergence and stability of learning. We draw on the research on {generative networks}~\cite{goodfellow2020generative} and neural network ensemble methods~\cite{wu2021boosting,chow2021robust,yanzhaowu2022boosting,ka2022object} to integrate explainable AI techniques into our safety enhanced AI algorithm design.

\begin{figure}[t]
\includegraphics[width=\linewidth]{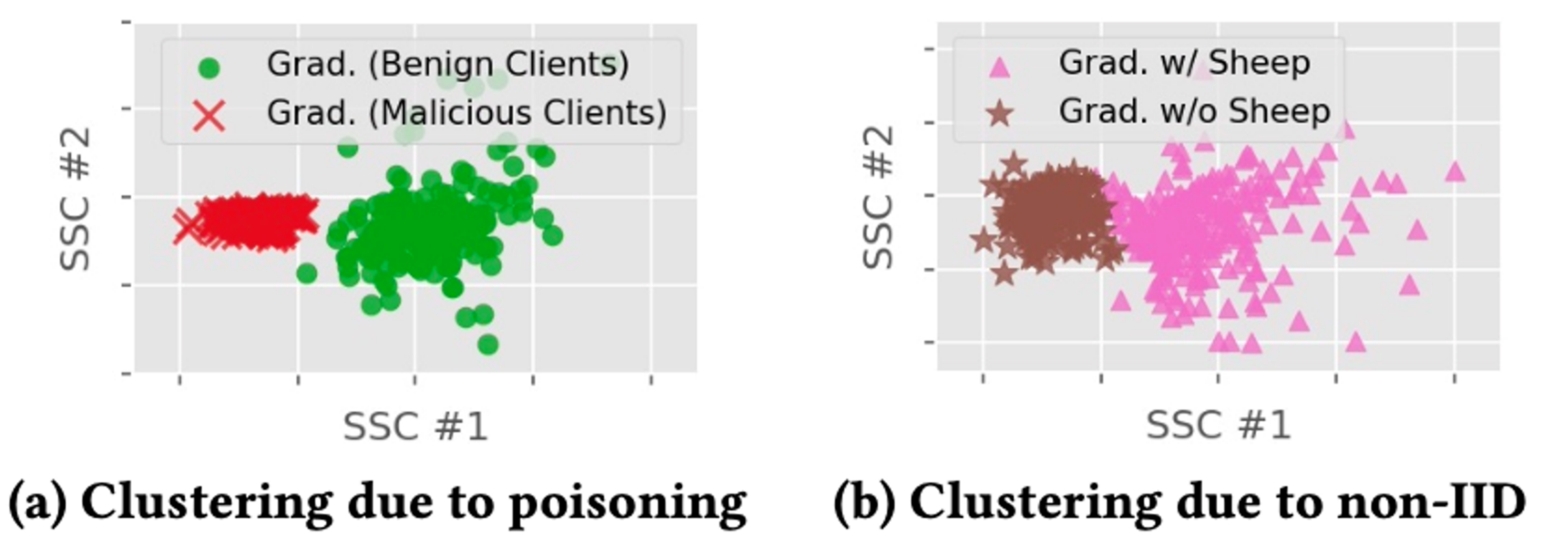}
\caption{{\small The non-IID scenario in Federated Learning may give a false sense of Trojan attack.} 
}\label{non-iid-trojan}
\end{figure}

Trojan attacks to AI models can cause incorrect monitoring, delivery, adversarial manipulation of pricing, abuse and misuse of DER management. 
Recent defense against Trojan attacks are based on a key observation that the clean samples of a certain class and the poisoned samples under Trojan attack to that class are disjoint in the problem space. 
Consequently, a mixture of two subgroups can be observed, one represents poisoned gradients and the other corresponds to non-poisoned gradients~\cite{chen2019detecting, tolpegin2020data, tran2018spectral, tang2021demon, hayase2021spectre}. 
The anomaly triggered by the trojaned samples can be characterized by the inconsistencies between the explainability representations and their predicted labels. This robust statistics based observation provides a basis for investigating the problem of characterizing trojaned samples from the perspective of out of distribution (OOD) detection~\cite{wei2020robust}.

\begin{figure}[t]
\includegraphics[width=\linewidth]{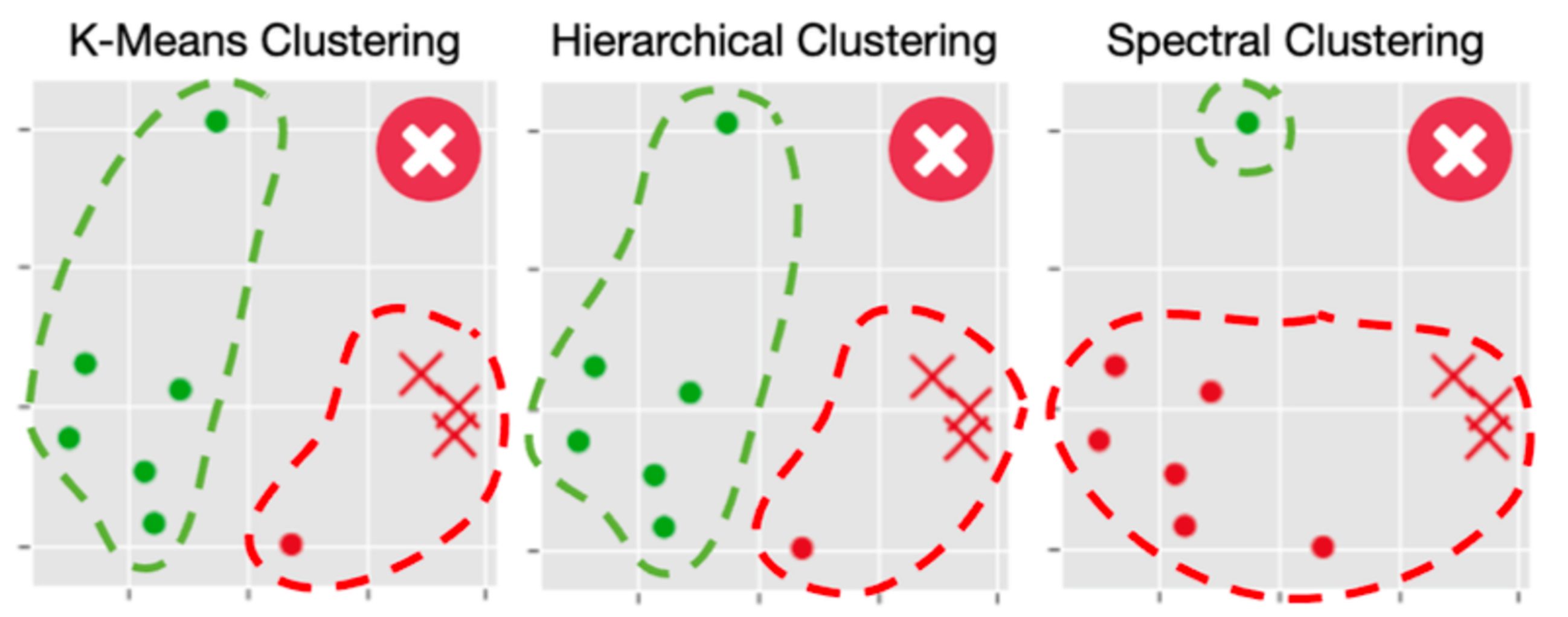}
\caption{{\small PCA followed by clustering fails to learn two clean clusters. Dots represent non-poisoned local model updates from benign clients, while crosses represent poisoned local model update from Trojan compromised clients.} }  
\label{fig_ssc}
\end{figure}

\begin{figure}[t]
\centering
\includegraphics[width=\linewidth]{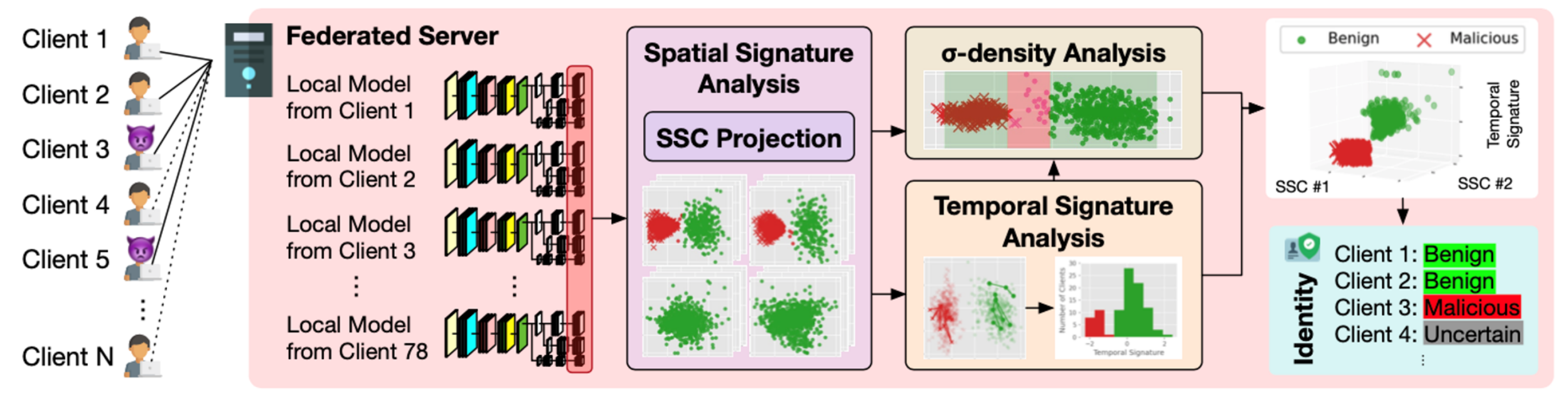}
\caption{{\small A three-tier explainable AI framework for Trojan detection and mitigation.} }
\label{fig-three-tier}
\end{figure}

However, defenses based on such robust statistics~\cite{tran2018spectral} suffer from two inherent limitations: (i) the non-iid scenario may give a false sense of Trojan attack (see Figure~\ref{non-iid-trojan}); and (ii) even with a small percentage of compromised clients, there may not be a clean separation of poisoned gradients due to Trojan attack at the compromised clients and non-poisoned gradients from benign clients (see Figure~\ref{fig_ssc}). In this solution, we propose a three tier explainable AI methodology for conducting spatiotemporal forensic analysis by creating spatial signature analysis, temporal signature analysis, and $\sigma$-density analysis (see Figure~\ref{fig-three-tier}). In contrast to previous works~\cite{tang2021demon, hayase2021spectre} assuming that the two subgroups follow Gaussian distributions with two different means but the same covariance, we employ Regularized Maximum Mean Discrepancy (RMMD)~\cite{danafar2013testing} to enhance the adversarial robustness of our method. RMMD is a Kernel-based two-sample testing technique which does not have any assumption on the distributions. In addition, we utilize Median Absolute Deviation (MAD), a more robust estimation of statistical dispersion, to measure the deviation of trojaned samples. 

\subsection{Responsible and Equitable AI Software Testing and Compliance Evaluation}
The third solution will develop responsible and equitable AI evaluation and compliance methods by integrating software testing and verification methods with explainable AI techniques. We develop compliance test metrics and evaluation platforms, including the Explainable AI guided feature selection techniques, the explainable AI model compliance measures (\eg, correctness and robustness under disruptive events), and the AI fairness testing tool to detect and remove biases in data collection and data interpretation. This solution also includes a proof of concept prototype development, aiming to demonstrate a selection of DER use-cases. 


We conduct both static and dynamic testings for fairness evaluation, focusing on DER domain specific fairness bugs. 
In static testings, we use historical or generated data to perform testing to check whether the system meets the fairness requirements defined using our fairness aware data collection framework and our policy-enhanced data governance guidelines (recall Section~\ref{sec_data_governance}). 
However, considering that test data cannot fully represent unseen or future data, static testing has its inherent limitation and it tends to miss those hidden circumstances that are problematic, for example, w.r.t. DER-specific fairness requirements, such as equitable access and bidding for energy consumption as well as fair energy allocation and supply.  
Dynamic testing employs run-time monitoring to keep tracking of the performance stability of an AI model, and continuously evaluating whether the AI software can still yield fair prediction outputs from the input test data~\cite{yang2021biasrv,tamburrelli2014towards}. During dynamic testing process, user feedback is a common source of feedbacks to determine the existence and type of fairness bugs. 
Specifically, we first identify and generate sound, complete, and correct test oracle for fairness testing, where the challenge is to obtain the ``ground truth'' of the input data, which is manually determined with DER domain-specific knowledge. Next, we leverage some existing test input generation techniques~\cite{zhang2020white,zhang2021efficient,zhang2021automatic}, which to date are mainly based on adversarial perturbation of input features. Hence, this approach cannot guarantee that the generated instances are legitimate and natural. In this solution, we investigate on methods to generate legitimate and natural test inputs for fairness testing, and the techniques to evaluate the naturalness of the generated test inputs, such as data distribution similarity analysis. In addition, we plan to design specific test selection, prioritization, and minimization techniques to reduce the cost of fairness testing without influencing the test effectiveness. 


To evaluate the responsible level of an AI model produced by our distributed learning algorithms, we conduct verification tests on inputs and outputs of the AI model, \eg, to identify the valid test inputs that satisfy the distribution closeness to that of the training
data. 
AI algorithms often suffer from overconfidence issues, which output high confidence probabilities on erroneous predictions, incl. adversarial or trojaned inputs. 
We investigate a set of complimentary quantified predictive uncertainty metrics to identify such responsible AI models misbehavior, including \textit{Bayesian-based metrics}: variation ratio
variation ratio of original prediction, prediction entropy, predictive entropy with dropout-enable, mutual information, and \textit{non-Bayesian metrics}: prediction confidence score, evidence-based metrics, ensemble-based metrics 
(prediction variance of ensemble models, and prediction variance of dropout-enable ensemble models).

\section{Conclusion}

In conclusion, this study has provided a comprehensive exploration of the dynamic landscape at responsible AI for enabling the Internet of Energy. By reviewing the overarching vision, the study has underscored the transformative potential of combining these domains to revolutionize energy management paradigms. Through a meticulous analysis of future challenges, ranging from data security to algorithmic transparency, the study has highlighted the complexity of the task at hand. However, the innovative solutions proposed herein offer a promising path forward. By emphasizing the importance of data governance, responsible algorithmic design, and  system architectures, this study provides actionable strategies that can guide stakeholders in realizing the full potential of this convergence. As we navigate the intricate interplay of technology and sustainability, these insights serve as a compass, directing us towards a future where the Internet of Energy and responsible AI empower each other for the greater good.

\section*{Acknowledgment}
This work is supported by the CSIRO (Australia) – National Science Foundation (US) AI Research Collaboration Program. 

\bibliographystyle{IEEEtranS}
\bibliography{ref}

\end{document}